\documentclass[10pt,twocolumn,letterpaper]{article}
\usepackage{meta/cvpr}              

\usepackage[dvipsnames]{xcolor}

\definecolor{cvprblue}{rgb}{0.21,0.49,0.74}
\usepackage[pagebackref,breaklinks,colorlinks,citecolor=cvprblue]{hyperref}

\usepackage{multirow}
\usepackage{algorithm}
\usepackage{algpseudocode}
\usepackage{pifont}
\usepackage{colortbl}
\usepackage{xcolor}
\newcommand{\xmark}{\ding{55}}
\newcommand{\cmark}{\ding{51}}
\newcommand{\et}{\textit{et al.\ }}

\definecolor{lightorange}{rgb}{0.8, 0.4, 0.0}  

\definecolor{darkgray}{rgb}{0.8, 0.8, 0.8}  
\algrenewcommand\algorithmicrequire{\textbf{Input:}}
\algrenewcommand\algorithmicensure{\textbf{Output:}}
\usepackage{arydshln}

\makeatletter
\renewcommand\@makefnmark{%
  \hbox{\textsuperscript{\normalfont\color{black}\@thefnmark}}%
}
\makeatother
\renewcommand{\thefootnote}{$\dagger$}

\def\confName{CVPR}
\def\confYear{2026}
\title{ActivityForensics: A Comprehensive Benchmark\\
for Localizing Manipulated Activity in Videos
\vspace{-0.7cm}
}

\author{
Peijun Bao$^{1,2}$,
~Anwei Luo$^{3,4\dagger}$,
~Gang Pan$^1$,
~Alex C. Kot$^{5,6,2}$,
~Xudong Jiang$^2$
\\
$^1$College of Computer Science and Technology, Zhejiang University
\\
$^2$School of Electrical and Electronic Engineering, Nanyang Technological University
\\
$^3$School of Computing and Artificial Intelligence, Jiangxi University of Finance and Economics
\\
$^4$Jiangxi Provincial Key Laboratory of Multimedia Intelligent Processing
\\
$^5$Faculty of Engineering, Shenzhen MSU-BIT University
\quad
$^6$VinUniversity
\\
peijun001@e.ntu.edu.sg \quad luoanwei@jxufe.edu.cn
\vspace{-.4cm}
}

\begin{document}
\maketitle
\newcommand\blfootnote[1]{%
  \begingroup
  \renewcommand\thefootnote{}\footnote{#1}%
  \addtocounter{footnote}{-1}%
  \endgroup
}


\begin{abstract}
\vspace{-0.2cm}
Temporal forgery localization aims to temporally identify manipulated segments in videos.
Most existing benchmarks focus on \textbf{appearance-level} forgeries, such as face swapping and object removal.
However, recent advances in video generation have driven the emergence of  \textbf{activity-level} forgeries that modify human actions to distort event semantics, resulting in highly deceptive forgeries that critically undermine media authenticity and public trust.
To overcome this issue, we introduce \textbf{ActivityForensics}, the first large-scale benchmark  for localizing manipulated activity in  videos.
It contains over 6K forged video segments that are seamlessly blended into the video context, rendering high visual consistency that makes them almost indistinguishable from authentic content to the human eye.
We further propose Temporal Artifact Diffuser (TADiff), a simple yet effective baseline that exposes artifact cues through a diffusion-based feature regularizer.
Based on ActivityForensics, we introduce comprehensive evaluation protocols covering intra-domain, cross-domain, and open-world settings, and benchmark a wide range of state-of-the-art forgery localizers to facilitate future research.
The dataset and code are available at \href{https://activityforensics.github.io}{https://activityforensics.github.io}.
%
\vspace{-0.3cm}
\end{abstract}
\blfootnote{$\dagger$: Corresponding author.}

\begin{figure}[t!]
\centering
\includegraphics[width=0.9\linewidth]{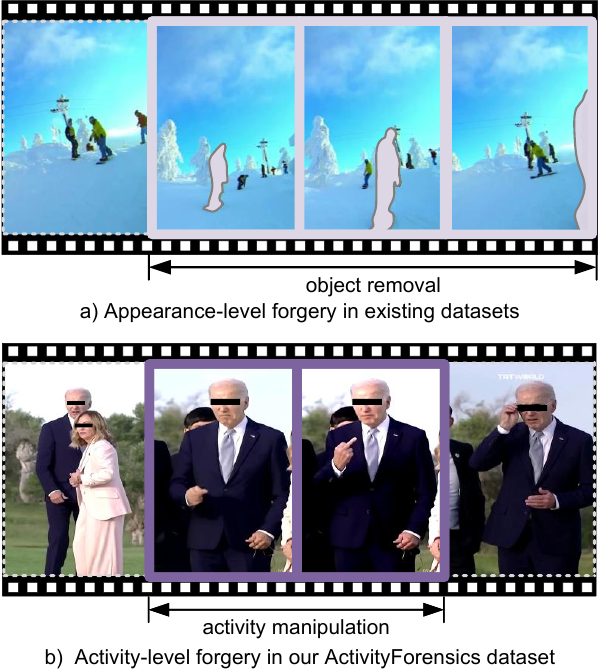}\vspace{0.15cm}
\scalebox{0.83}{
\begin{tabular}{ccc}
\toprule
Forgery level & Dataset & Forgery types \\
\midrule
\multirow{4}{*}{Appearance-level} & ForgeryNet~\cite{He2021ForgeryNet} & Face \\
& Lav-DF~\cite{Cai2022DoYouReallyMeanThat} & Face \\
& AV-Deepfake1M~\cite{Cai20241MDeepfakes} & Face \\
& TVIL~\cite{Zhang2023UMMAFormer} & Object Removal \\
\midrule
\textbf{Activity-level} & \textbf{ActivityForensics (Ours)} & \textbf{Activity} \\
\bottomrule
\end{tabular}
}
\par\vspace{0.1cm}
\centering 
c) Comparison to temporal forgery localization datasets
\vspace{-0.2cm}
\caption{
a)~Existing datasets for temporal forgery localization mainly focus on \textbf{appearance-level} forgeries such as object removal and face manipulation.
b)~Driven by the remarkable advances in video generation and editing in recent years, however, \textbf{activity-level} forgeries have become increasingly prevalent and pose significant risks to media integrity and societal trust.
c)~To address this emerging threat, we present ActivityForensics, the first dataset for localizing manipulated activities in videos.
}
\vspace{-0.6cm}
\label{fig_motivation}
\end{figure}

\section{Introduction}
\label{sec_intro}
%
%
With the rapid advancement of generative and editing technologies, the creation of highly realistic yet falsified video content has become increasingly accessible~\cite{Zhu2025GenerativeIT,Chen2025SciFi,Jiang2025VACEAV,Chen2025EFVI,Wang2025Wan,Xing2024Survey,Sun2024Diffusion,Luo2023VideoFusion,HaCohen2024LTXVideo}.
Sophisticated deep learning models now enable seamless synthesis, replacement, or alteration of visual elements in videos, often yielding manipulated content that is nearly indistinguishable from authentic footage.
This growing capability has raised serious concerns about misinformation and the integrity of multimedia evidence.
As a result, developing reliable methods for localizing video forgery~\cite{Rossler2019FaceForensics,Pei2024Deepfake,Javed2021ACS,Yu2021ASO} has emerged as a critical research direction in multimedia forensics and trustworthy artificial intelligence.
As shown in Fig.~\ref{fig_motivation}, existing benchmarks for temporal forgery localization mainly focus on \textbf{appearance-level} forgery such as face manipulation~\cite{He2021ForgeryNet,Cai2022DoYouReallyMeanThat,Cai20241MDeepfakes} and object removal~\cite{Zhang2023UMMAFormer}.
However, due to significant progress in video generation and editing in recent years, \textbf{activity-level} forgeries have become increasingly common in social media and video platforms.
Fig.~\ref{fig_motivation} b) illustrates a representative example taken from a news video featuring a politician at a diplomatic event: within an otherwise authentic stream, a brief segment is subtly manipulated so that a neutral standing posture is transformed into a gesture of misconduct.
Such manipulation is coherently blended into the rest of the video, making the manipulation boundaries subtle and resulting in highly deceptive forgeries that critically undermine media authenticity and public trust~\cite{Shahbazi2024Social,VanZoonen2024Trust}.

To fill this gap, we introduce \textbf{ActivityForensics}, the first large-scale dataset specifically designed for manipulated activity localization in videos.
A key challenge in collecting such a dataset is the labor-intensive manual effort to select appropriate video segments and smoothly embed manipulated ones into neighboring content.
To overcome this, we propose grounding-assisted data construction that automatically inserts manipulated activity segments into appropriate video contexts and produces precise temporal annotations without human intervention.
Specifically, we leverage video captioning and grounding~\cite{dense_cap,Bao2021Dense,gao2017tall} to obtain activity descriptions and localize their corresponding temporal segments.
These descriptions are subsequently manipulated to create semantically altered counterparts via Large Language Models (LLMs)~\cite{OpenAI2023GPT4}.
Finally, we condition video generation and editing models~\cite{Wang2025Wan,Zhu2025GenerativeIT,Chen2025SciFi,Jiang2025VACEAV,HaCohen2024LTXVideo} on both the manipulated descriptions and the grounding information to synthesize activity-level forgeries.
In this way, the manipulated segments are seamlessly integrated into the original video contexts, achieving a high level of visual and temporal realism that makes them difficult for human observers to distinguish from authentic content.

Alongside the dataset, we further establish three evaluation settings, namely intra-domain, cross-domain, and open-world settings to systematically assess performance across diverse manipulation domains.
We conduct extensive benchmarking for manipulated activity localization with a broad spectrum of state-of-the-art approaches~\cite{Zhang2022ActionFormer,Zhang2023UMMAFormer,Kim2025DiGIT} adapted from temporal action localization and temporal forgery localization.
While most temporal forgery localization models adopt architectures inherited from action localization, the two tasks differ fundamentally: action localization relies on high-level semantics for event understanding, whereas manipulated activity localization requires sensitivity to subtle temporal and visual artifacts.
To this end, we propose Temporal Artifact Diffuser (TADiff), a simple yet effective baseline that injects stochastic perturbations into the multi-scale feature space to mitigate semantic bias and progressively denoises them to amplify subtle forgery-discriminative signals.

In summary, our contributions are threefold:
\begin{itemize}
\item We propose a new task of manipulated activity localization and introduce the first large-scale dataset tailored for it.
A grounding-assisted framework is devised to harmoniously embed manipulated segments into the surrounding footage, facilitating scalable dataset construction with precise temporal annotations.
\item Alongside the dataset, we introduce extensive evaluation protocols covering intra-domain, cross-domain, and open-world settings, and provide extensive benchmarks of state-of-the-art approaches on this new task.
\item A Temporal Artifact Diffuser (TADiff) is proposed to effectively capture forgery evidence through a diffusion-based feature regularizer.
\end{itemize}
We believe ActivityForensics will serve as a cornerstone for advancing fine-grained video forensics research and fostering digital integrity infrastructures.

\begin{figure*}[t!]
\centering
\includegraphics[width=0.96\linewidth]{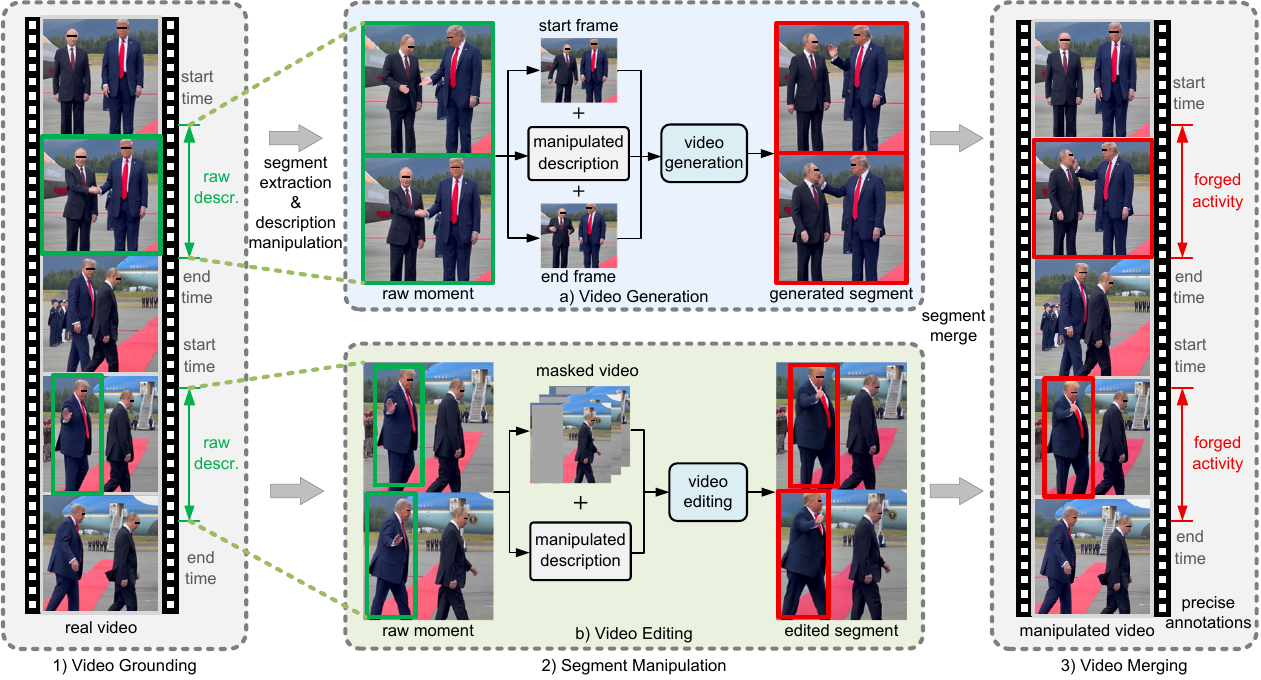}
\vspace{-0.1cm}
\caption{
Overview of grounding-assisted data generation pipeline.
1)~We leverage video captioning and temporal grounding to obtain activity descriptions and localize their corresponding temporal segments.
2)~Subsequently, grounded segments and manipulated descriptions are harnessed as conditioning signals to automatically perform activity manipulations.
3)~The manipulated segments are finally seamlessly merged into the rest of the video, while remaining visually consistent across both tampered and authentic regions.
The green bounding boxes indicate the original regions, while the red ones correspond to the manipulated regions.
}\label{fig_generation}
\vspace{-.3cm}
\end{figure*}

\section{Related Works}
\subsection{Video Manipulation Methods}
Recent advances in video manipulation are largely driven by conditioned video generation and masked video editing.
For conditioned generation methods, models such as Wan~\cite{Wang2025Wan}, FCVG~\cite{Zhu2025GenerativeIT}, Scifi~\cite{Chen2025SciFi}, and Vidu~\cite{Bao2024Vidu} synthesize temporally coherent sequences under text, pose, or key-frame conditioning, enabling controllable and high-fidelity creation of new actions.
For masked video editing, approaches including the VACE framework~\cite{Jiang2025VACEAV} and LTX~\cite{HaCohen2024LTXVideo} perform localized modifications guided by prompts, masks, and frame constraints while preserving the surrounding appearance and motion.
The realism and controllability offered by these generation and editing techniques make manipulated activities increasingly seamless and deceptive, thereby heightening both the technical challenges and societal risks associated with video forgery~\cite{kong2026open,zou2025bi,zou2025semantic}.

\subsection{Temporal Forgery Localization}
The increasing accessibility of video manipulation techniques has raised significant concerns regarding media authenticity~\cite{Shahbazi2024Social,VanZoonen2024Trust}.
As real-world manipulation typically occurs within short temporal moments in untrimmed videos,  temporal forgery localization has become a fundamental problem in video forensics~\cite{Rossler2019FaceForensics,Pei2024Deepfake,Javed2021ACS,Yu2021ASO}.
%
Zhang~\et~\cite{Zhang2023UMMAFormer} propose a temporal video inpainting localization benchmark.
ForgeryNet~\cite{He2021ForgeryNet}, Lav-DF~\cite{Cai2022DoYouReallyMeanThat}, and AV-Deepfake1M~\cite{Cai20241MDeepfakes} are representative works for temporal localization of face manipulation.
Unlike these previous works that focus on appearance-level forgery, we are the first to study the localization of activity-level manipulation.

\subsection{Temporal Video Localization}
Localizing temporal moments of interest in videos has recently received increasing attention.
%
The tasks most closely related to ours include temporal action localization~\cite{Zhang2023HOIAware,bao2023cross,Shou2016Temporal,Liberatori2024TestTime,Zhang2022ActionFormer}, temporal grounding~\cite{Dong2024GraphBased,bao2024e3m,Wu2025Number,bao2024local}, and video anomaly detection~\cite{Saligrama2012Video,Zhou2024Video,Zanella2024Harnessing}.
Specifically, temporal action localization~\cite{Shou2016Temporal} aims to identify and temporally localize specific actions within untrimmed videos.
Video grounding extends this idea by localizing video moments described by language queries, and recent works~\cite{Yang2023AttractiveSS,Yang2024SynchronizedVS,bao2024e3m} have achieved significant progress through effective multimodal alignment.
Video anomaly detection~\cite{Saligrama2012Video}, on the other hand, focuses on identifying semantically abnormal events such as fighting or explosions.
Distinct from these tasks, which require understanding high-level event semantics, our goal is to identify the temporal moments during which manipulated activities occur, relying on subtle visual inconsistencies rather than semantic cues.

\begin{figure*}[t!]
\centering
\scalebox{1.1}{
    \begin{subfigure}[t]{0.36\linewidth}
        \centering
        \includegraphics[width=\linewidth]{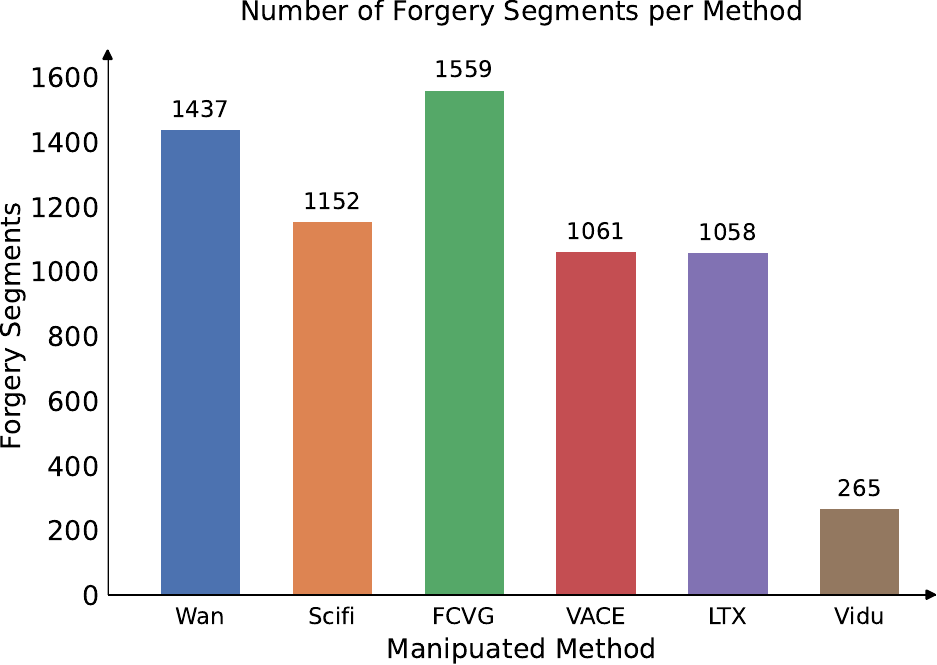}
        \caption{}
        \label{fig_sub1}
    \end{subfigure}
    \hfill
    \begin{subfigure}[t]{0.285\linewidth}
        \centering
        \includegraphics[width=\linewidth]{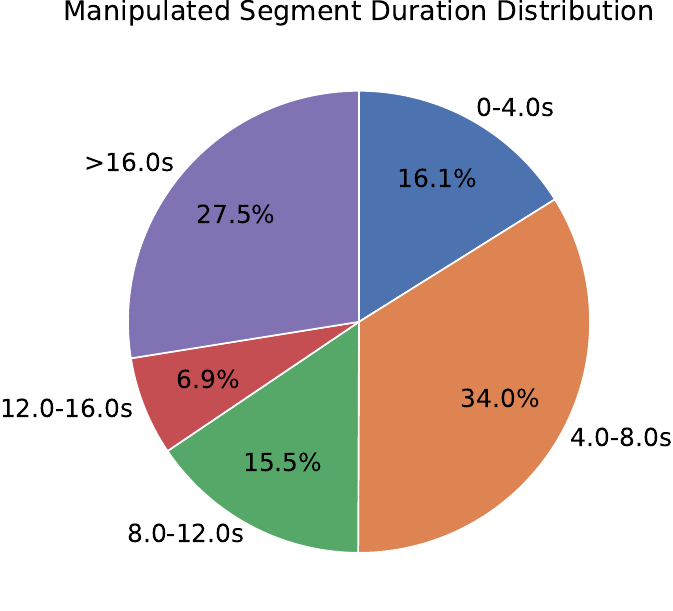}
        \caption{}
        \label{fig_sub2}
    \end{subfigure}
    \hfill
    \begin{subfigure}[t]{0.285\linewidth}
        \centering
        \includegraphics[width=\linewidth]{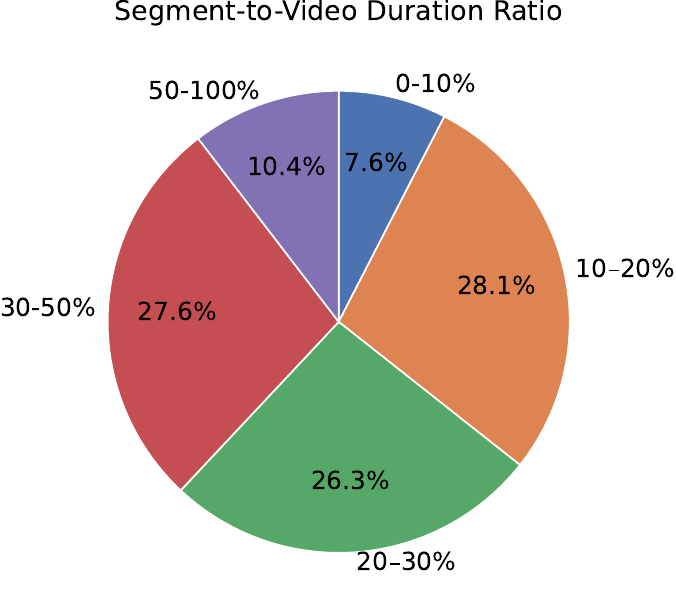}
        \caption{}
        \label{fig_sub3}
    \end{subfigure}
}
\vspace{-0.2cm}
\caption{
Statistics of the ActivityForensics dataset.
a) Histogram of forgery-segment counts across manipulation methods, where vidu is used only for evaluation.
b) Distribution of manipulated segment durations.
c) Distribution of the ratio between manipulated segment duration and overall video duration.
}
\label{fig_data_stat_fake_seg}
\vspace{-0.4cm}
\end{figure*}

\section{ActivityForensics}\label{sec_dataset}

\subsection{Grounding-Assisted Data Construction}
In real-world scenarios, activity manipulation typically demands extensive manual effort to carefully select appropriate video segments and then smoothly embed manipulated ones into neighboring content to avoid noticeable visual or temporal discontinuities.
However, such manual construction is time-consuming and impractical at scale.
To tackle this challenge, as illustrated in Fig~\ref{fig_generation}, we propose grounding-assisted data construction, which leverages video captioning and grounding to coherently embed manipulated segments into video contexts without manual intervention and produce precise temporal annotations.
Specifically, 1)~we first exploit video captioning and temporal grounding~\cite{dense_cap,bao2024local} to obtain activity descriptions and localize their corresponding temporal segments.
We then manipulate the original descriptions to create semantically altered counterparts using large language models~\cite{OpenAI2023GPT4}.
For instance, the original description ``the man waves his hands'' in Fig~\ref{fig_generation} is transformed to ``the man gives a thumbs-up''.
2)~Subsequently, we apply video manipulation methods to synthesize activity-level forgeries with high visual fidelity.
We consider two typical categories of manipulation models: \textit{video generation} models~\cite{Wang2025Wan,Zhu2025GenerativeIT,Chen2025SciFi} that synthesize all frames within the manipulated segment
and \textit{video editing} models~\cite{Jiang2025VACEAV,HaCohen2024LTXVideo} that modify only the masked region of the video segment while preserving the background content.
Both the manipulated descriptions and the grounding information such as start and end frames are exploited as conditioning signals for generation or editing, thereby producing segments that naturally align with the surrounding video content.
3)~Finally, we replace the original segments with the synthesized ones and reintegrate them into the video, achieving high visual and temporal realism that makes the manipulations difficult for human observers to distinguish from authentic content.
More details on data construction, including the video sources, LLM prompting strategy, and the human evaluation of data quality, are provided in the supplementary material.

\begin{table}[t!]
\caption{
Summary of manipulation methods in ActivityForensics.
}\label{table_manipulation}
\scalebox{0.95}{
\begin{tabular}{cllll}
\toprule
Category & Method & Guidance type\\
\midrule
\multirow{4}{*}{Video Generation}
&
Wan~\cite{Wang2025Wan}  & Text driven
\\
&
Scifi~\cite{Chen2025SciFi} & Frame Interpolation
\\
&
FCVG~\cite{Zhu2025GenerativeIT} & Pose driven
&
\\
&
Vidu~\cite{Bao2024Vidu} & Commercial API
\\
\midrule
\multirow{2}{*}{Video Editing}
&
VACE~\cite{Jiang2025VACEAV} & Text driven
\\
&
LTX~\cite{HaCohen2024LTXVideo} & Text driven
\\
\bottomrule
\end{tabular}
}
\vspace{-0.5cm}
\end{table}

\subsection{Dataset Statistics}
Table~\ref{table_manipulation} summarizes the manipulation methods used in ActivityForensics, grouped into two major categories:
\textit{video generation} models, including Wan~\cite{Wang2025Wan}, Scifi~\cite{Chen2025SciFi}, FCVG~\cite{Zhu2025GenerativeIT}, and the commercial system Vidu~\cite{Bao2024Vidu}, and \textit{video editing} models, including VACE~\cite{Jiang2025VACEAV} and LTX~\cite{HaCohen2024LTXVideo}.
These methods collectively span key forgery paradigms such as text-driven generation, pose-driven motion synthesis, and region-constrained editing.
We do not include other video generative models such as Sora~\cite{openai2024sora}, as they do not support controlled start–end frame conditioning, and their generated segments cannot be well aligned with the rest of video.
Fig.~\ref{fig_sub1} further presents the number of forgery segments for each manipulation method, with vidu included only for testing.
The dataset contains over 6,000 forgery segments, distributed evenly across different manipulation mechanisms to ensure balanced coverage.
As shown in Fig.~\ref{fig_sub2}, the durations of manipulated segments vary widely, providing a rich and diverse distribution.
Moreover, Fig.~\ref{fig_sub3} illustrates the distribution of the ratio between manipulated-segment duration and overall video duration. 
More than $60\%$ of manipulated segments occupy less than $30\%$ of the corresponding video, highlighting the challenge of accurately localizing them.
Additional dataset statistics can be found in the supplementary material.

\subsection{Temporal Artifact Diffuser}
\textbf{Problem Formulation.}
The goal of manipulated activity localization is to identify forged segments in long, untrimmed videos by predicting the temporal intervals that contain manipulated activities.
Formally, given a video $V = \{v_t\}_{t=1}^T$, the task is to predict a set of temporal intervals $\{(\tau_s, \tau_e)\}$, each corresponding to a manipulated segment within the video.

\begin{figure*}[t!]
\centering
\includegraphics[width=0.95\linewidth]{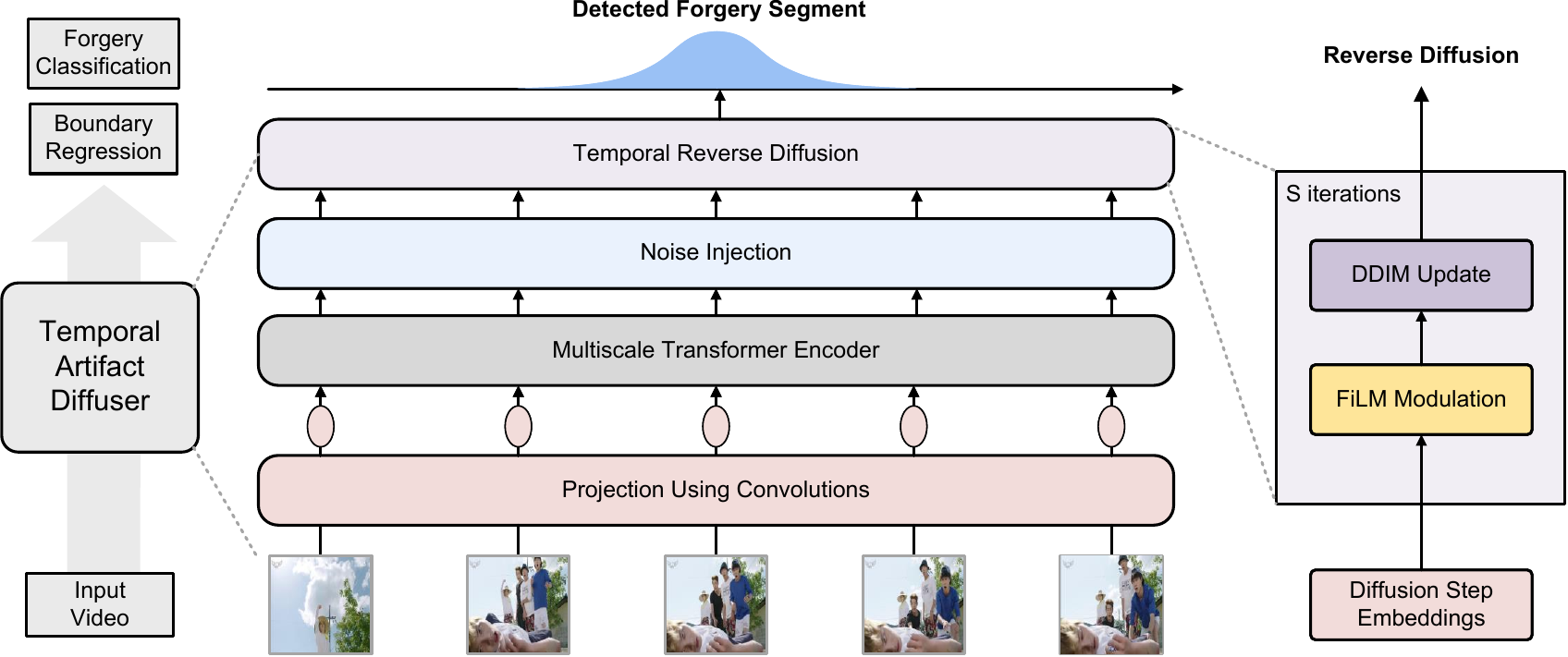}
\caption{
Overview of Temporal Artifact Diffuser (TADiff).
Different from action localization that relies on high-level semantics for event understanding, manipulated activity localization requires sensitivity to subtle temporal and visual artifacts.
To this end, TADiff injects stochastic perturbations into the temporal feature space of ActionFormer to suppress semantic bias, and then amplifies artifact cues via iterative denoising, composed of Feature-wise Linear Modulation (FiLM) and Denoising Diffusion Implicit Model (DDIM) updates.
}\label{fig_method}
\end{figure*}

\noindent
\textbf{Motivations.}
Model architectures originally developed for temporal action localization are widely adopted in the area of temporal forgery localization.
However, unlike action localization which depends on high-level semantics such as event type, forgery localization relies on subtle low-level cues that are largely independent of semantics, including texture irregularities and motion discontinuities.
As a result, models directly adapted from temporal action localization often overfit to \textit{semantic bias}, limiting their generalization in manipulated activity localization.
To overcome this, we propose a simple yet effective diffusion-based feature regularization dubbed \textit{Temporal Artifact Diffuser (TADiff)}.
TADiff injects stochastic perturbations into the temporal feature space to suppress semantic bias, and then amplifies forgery-discriminative signals via an iterative denoising process consisting of  Feature-wise Linear Modulation (FiLM) and Denoising Diffusion Implicit Model (DDIM) updates.
This process effectively regularizes the feature manifold, discourages over-reliance on semantics, and improves sensitivity to subtle artifact cues critical for manipulated activity localization.

\noindent
\textbf{Model Architecture.}
Given the frame-level embeddings $X = \{x_t\}_{t=1}^T \in \mathbb{R}^{T \times C}$ extracted from a visual backbone, we follow ActionFormer~\cite{Zhang2022ActionFormer} to build a temporal feature pyramid with a multi-scale transformer encoder.
The pyramid aggregates contextual information at multiple temporal resolutions, producing feature sequences
\begin{equation}
f^{(l)} \in \mathbb{R}^{N_l \times C}, \quad l = 1, \dots, L,
\end{equation}
where $N_l$ denotes the temporal length at level $l$ and $C$ is the shared feature dimension.
Each temporal location in $f^{(l)}$ captures local temporal context that may correspond to either authentic or forged content.
However, the representations in action-localization architectures are primarily shaped by high-level semantics.
While informative for action understanding, these cues contribute little to forgery discrimination, which limits the model's ability to generalize across manipulation types.

To alleviate this issue, we introduce TADiff after the multi-scale Transformer network to regularize and refine temporal features before prediction.
TADiff operates as a deterministic denoising chain that explicitly models both forward noise injection and reverse denoising of temporal representations, encouraging the network to learn artifact-sensitive and semantically invariant features.
For simplicity, we describe the process for one temporal feature sequence $f \in \mathbb{R}^{N \times C}$.
In the forward process, Gaussian noise is added to the feature sequence:
\begin{equation}
x_s = \sqrt{\bar{\alpha}_s}f + \sqrt{1 - \bar{\alpha}_s}\epsilon, \quad \epsilon \sim \mathcal{N}(0, I),
\end{equation}
where $\bar{\alpha}_s$ follows a linear noise schedule that determines the perturbation strength.
This step perturbs the representation away from its semantic manifold and introduces stochasticity into the temporal feature space.

After perturbation, the model performs denoising to augment forgery-discriminative representations.
This reverse process is parameterized by a lightweight temporal convolutional denoiser, implemented with Feature-wise Linear Modulation (FiLM)~\cite{Perez2017FiLM}, which predicts and removes the injected noise conditioned on the diffusion step $s$.
The model progressively reconstructs an artifact-sensitive signal through a deterministic reverse process inspired by Denoising Diffusion Implicit Models (DDIM)~\cite{Song2020DDIM}, formulated as:
\begin{equation}
x_{s-1} = \sqrt{\bar{\alpha}_{s-1}}\hat{x}_0 + \sqrt{1 - \bar{\alpha}_{s-1} - \sigma_s^2}\hat{\epsilon} + \sigma_s z,
\end{equation}
where $\hat{x}_0$ and $\hat{\epsilon}$ denote the predicted artifact-enhanced feature and residual noise, $z \sim \mathcal{N}(0, I)$ is a Gaussian perturbation, and $\sigma_s$ (controlled by coefficient $\eta$) defines the stepwise randomness.
Through this progressive denoising process, TADiff refines forgery-aware representations that complement the underlying semantic structure of the video.

\begin{table*}[t!]
\centering
\caption{
Quantitative comparisons under intra-domain and open-world settings.
Each section reports Average Precision (AP) at multiple tIoU thresholds and Average Recall (AR) at various proposal counts.
Orange numbers indicate improvements over the ActionFormer baseline on which our TADiff is built.
}
\label{table_inout_domain}
\vspace{-0.2cm}
\scalebox{0.81}{
\begin{tabular}{c l@{\hspace{-8pt}}
c@{\hspace{-17pt}} c@{\hspace{-17pt}}
c@{\hspace{-17pt}} c@{\hspace{-17pt}}
c@{\hspace{-17pt}} c@{\hspace{-17pt}}
c@{\hspace{-17pt}} c@{\hspace{0pt}}
}
\toprule
\multirow{2}{*}{Domain} &
\multirow{2}{*}{Method }
& \multicolumn{4}{c}{AP metrics}
& \multicolumn{4}{c}{AR metrics}
\\
\cmidrule(l{13pt}r{5pt}){3-6}
\cmidrule(l{15pt}r{12pt}){7-10}
& & AP@0.75 & AP@0.85 & AP@0.95 & avg
& AR@1 & AR@5 & AR@10 & avg
\\
\midrule
\multirow{4}{*}{Intra-Domain}
& ActionFormer~\cite{Zhang2022ActionFormer}~\textcolor{gray}{\scriptsize{ECCV22}}
& 86.29 & 78.92 & 46.79 & 70.67
& 63.85 & 78.80 & 80.28 & 74.31
\\
& UMMAFormer~\cite{Zhang2023UMMAFormer}~\textcolor{gray}{\scriptsize{MM23}}
& \underline{87.02} & \underline{80.25} & \underline{48.55} & \underline{71.94}
& \underline{64.75} & \underline{80.59} & \underline{81.88} & \underline{75.74}
\\
& DiGIT~\cite{Kim2025DiGIT}~\textcolor{gray}{\scriptsize{CVPR25}}
& 78.61 & 70.52 & 44.92 & 64.69
& 59.70 & 74.67 & 76.93 & 70.43
\\
& \textbf{TADiff (Ours)}
& \quad \quad \ \textbf{87.52}~{\scriptsize\textcolor{orange}{(+1.23)}}
& \quad \quad \ \textbf{81.05}~{\scriptsize\textcolor{orange}{(+2.13)}}
& \quad \quad \ \textbf{56.57}~{\scriptsize\textcolor{orange}{(+9.78)}}
& \quad \quad \ \textbf{75.05}~{\scriptsize\textcolor{orange}{(+4.38)}}
& \quad \quad \ \textbf{66.40}~{\scriptsize\textcolor{orange}{(+2.55)}}
& \quad \quad \ \textbf{81.84}~{\scriptsize\textcolor{orange}{(+3.04)}}
& \quad \quad \ \textbf{83.20}~{\scriptsize\textcolor{orange}{(+2.92)}}
& \quad \quad \ \textbf{77.15}~{\scriptsize\textcolor{orange}{(+2.84)}}
\\
\midrule
\multirow{4}{*}{Open-World}
& ActionFormer~\cite{Zhang2022ActionFormer}~\textcolor{gray}{\scriptsize{ECCV22}}
& 89.81 & 86.58 & 57.08 & 77.82 & 81.26 & 84.15 & 84.53 & 83.31
\\
& UMMAFormer~\cite{Zhang2023UMMAFormer}~\textcolor{gray}{\scriptsize{MM23}}
& \underline{91.13} & \underline{87.67} & \underline{57.57} & \underline{78.79}
& \underline{83.02} & \underline{84.65} & \underline{84.78} & \underline{84.15}
\\
& DiGIT~\cite{Kim2025DiGIT}~\textcolor{gray}{\scriptsize{CVPR25}}
& 88.99 & 79.93 & 55.26 & 74.73
& 77.74 & 80.88 & 81.38 & 80.00
\\
& \textbf{TADiff (Ours)}
& \quad \quad \ \textbf{92.35}~{\scriptsize\textcolor{orange}{(+2.54)}} & \quad \quad \ \textbf{89.52}~{\scriptsize\textcolor{orange}{(+2.94)}}
& \quad \quad \ \ \textbf{69.06}~{\scriptsize\textcolor{orange}{(+11.98)}}
& \quad \quad \ \textbf{83.64}~{\scriptsize\textcolor{orange}{(+5.82)}}
& \quad \quad \ \textbf{85.66}~{\scriptsize\textcolor{orange}{(+4.40)}}
& \quad \quad \ \textbf{88.68}~{\scriptsize\textcolor{orange}{(+4.53)}}
& \quad \quad \ \textbf{89.43}~{\scriptsize\textcolor{orange}{(+4.90)}}
& \quad \quad \ \textbf{87.92}~{\scriptsize\textcolor{orange}{(+4.61)}} \
\\
\bottomrule
\end{tabular}}
\end{table*}

\begin{table*}[t!]
\centering
\caption{
Quantitative comparisons under cross-domain scenarios.
Orange numbers illustrate gains over the ActionFormer baseline.
}
\vspace{-0.2cm}
\label{table_A_B}
\scalebox{0.85}{
\begin{tabular}{l l@{\hspace{-6pt}}
c@{\hspace{-17pt}} c@{\hspace{-17pt}}
c@{\hspace{-17pt}} c@{\hspace{-17pt}}
c@{\hspace{-17pt}} c@{\hspace{-17pt}}
c@{\hspace{-17pt}} c@{\hspace{0pt}}
}
\toprule
\multirow{2}{*}{Protocol} &
\multirow{2}{*}{Method }
& \multicolumn{4}{c}{AP metrics}
& \multicolumn{4}{c}{AR metrics}
\\
\cmidrule(l{13pt}r{5pt}){3-6}
\cmidrule(l{16pt}r{10pt}){7-10}
& & AP@0.75 & AP@0.85 & AP@0.95 & avg
& AR@1 & AR@5 & AR@10 & avg
\\
\midrule

\multirow{4}{*}{$A \rightarrow B$ }
& ActionFormer~\cite{Zhang2022ActionFormer}~\textcolor{gray}{\scriptsize{ECCV22}}
&85.61 & 76.73 & \underline{39.20} & 67.18 & 62.93 & 75.99 & 77.51 & 72.14
\\
& UMMAFormer~\cite{Zhang2023UMMAFormer}~\textcolor{gray}{\scriptsize{MM23}}
&\underline{88.42} & \underline{79.89} & 36.83 & \underline{68.38} & \underline{64.88} & \underline{78.92} & \underline{80.22} & \underline{74.67}
\\
& DiGIT~\cite{Kim2025DiGIT}~\textcolor{gray}{\scriptsize{CVPR25}}
&81.58 & 69.94 & 36.28 & 62.60 & 60.65 & 72.85 & 74.42 & 69.31
\\
& \textbf{TADiff (Ours)}
& \textbf{\quad \quad  \ 88.73}~{\scriptsize\textcolor{orange}{(+3.12)}}
& \textbf{\quad \quad  \ 80.36}~{\scriptsize\textcolor{orange}{(+3.63)}}
& \textbf{\quad \quad  \ 39.81}~{\scriptsize\textcolor{orange}{(+0.61)}}
& \textbf{\quad \quad  \ 69.63}~{\scriptsize\textcolor{orange}{(+2.45)}}
& \quad \quad  \ \textbf{65.04}~{\scriptsize\textcolor{orange}{(+2.11)}}
& \textbf{\quad \quad  \ 79.35}~{\scriptsize\textcolor{orange}{(+3.36)}}
& \textbf{\quad \quad  \ 80.33}~{\scriptsize\textcolor{orange}{(+2.82)}}
& \quad \quad  \ \textbf{74.91}~{\scriptsize\textcolor{orange}{(+2.77)}}
\\
\midrule
\multirow{4}{*}{$B \rightarrow A$}
& ActionFormer~\cite{Zhang2022ActionFormer}~\textcolor{gray}{\scriptsize{ECCV22}}
&55.57 & 42.31 & 13.54 & 37.14 & 40.76 & \underline{55.10} & \underline{57.22} & \underline{51.03}
\\
& UMMAFormer~\cite{Zhang2023UMMAFormer}~\textcolor{gray}{\scriptsize{MM23}}
&\underline{57.52} & \underline{44.35} & 12.90 & \underline{38.26} & \underline{40.85} & 54.64 & 56.63 & 50.71
\\
& DiGIT~\cite{Kim2025DiGIT}~\textcolor{gray}{\scriptsize{CVPR25}}
&45.49 & 33.50 & \underline{13.75} & 30.91 & 32.47 & 48.94 & 53.41 & 44.94
\\
& \textbf{TADiff (Ours)}
& \textbf{\quad \quad  \ 60.50}~{\scriptsize\textcolor{orange}{(+4.93)}}
& \textbf{\quad \quad  \ 47.64}~{\scriptsize\textcolor{orange}{(+5.33)}}
& \textbf{\quad \quad  \ 14.52}~{\scriptsize\textcolor{orange}{(+0.98)}}
& \textbf{\quad \quad  \ 40.89}~{\scriptsize\textcolor{orange}{(+3.75)}}
& \textbf{\quad \quad  \ 43.31}~{\scriptsize\textcolor{orange}{(+2.55)}}
& \textbf{\quad \quad  \ 56.23}~{\scriptsize\textcolor{orange}{(+1.13)}}
& \textbf{\quad \quad  \ 58.15}~{\scriptsize\textcolor{orange}{(+0.93)}}
& \textbf{\quad \quad  \ 52.56}~{\scriptsize\textcolor{orange}{(+1.53)}}
\\
\bottomrule
\end{tabular}}
\end{table*}

\noindent
\textbf{Objective Function.}
As our goal is to refine artifact-discriminative features rather than reconstruct the original content, the denoising process in TADiff is optimized solely under the localization objective.
Following ActionFormer~\cite{Zhang2022ActionFormer}, two prediction heads are applied at each temporal location in the multi-scale feature pyramid: a forgery confidence head estimating the likelihood of being a forged segment, and a boundary regression head predicting the offsets to its start and end boundaries.
The total training loss is defined as:
\begin{equation}
\mathcal{L} = \mathcal{L}_{cls} + \mathcal{L}_{reg},
\end{equation}
where $\mathcal{L}_{cls}$ is a focal loss on confidence scores, and $\mathcal{L}_{reg}$ is a smooth L1 loss for boundary regression.
TADiff can then be trained end-to-end to guide the diffusion dynamics to focus on temporal inconsistencies and subtle visual artifacts.

\section{Experiments}
\subsection{Implementation Details}
TADiff is built based on ActionFormer~\cite{Zhang2022ActionFormer}, which we use as the basic network architecture.
We train our model using the AdamW optimizer~\cite{Loshchilov2017DecoupledWD} with a batch size of $16$ and a learning rate of $0.001$.
The number of denoising steps is set to 3.
%
All other implementation details follow ActionFormer~\cite{Zhang2022ActionFormer} and are provided in the supplement.

\subsection{Benchmarking ActivityForensics}
\subsubsection{Benchmark Settings}
\noindent
\textbf{Evaluation Protocols.}
Our evaluation focuses on two key aspects: whether the model can achieve precise temporal localization under forgery distributions consistent with training, and whether it can maintain performance when tested on different forgery mechanisms, including previously unseen manipulation models.
To comprehensively examine these aspects, we conduct experiments under three evaluation settings:
\begin{itemize}
\item %
\textit{Intra-domain} setting: training and testing videos are manipulated by the same set of models, including Wan, Scifi, VACE, FCVG, and LTX.
\item
\textit{Open-world} setting: models are trained on manipulations from Wan, Scifi, VACE, FCVG, and LTX, and tested on unseen forgeries from the commercial model Vidu.
\item
\textit{Cross-domain} setting:
we define two transfer directions: A$\rightarrow$B and B$\rightarrow$A.
The A domain consists of video generation methods, including
Wan~(text-driven generation),
Scifi~(frame interpolation),
and VACE~(text-driven editing).
The B domain includes
FCVG~(pose-driven generation)
and LTX~(text-driven editing).
\end{itemize}

\noindent
\textbf{Evaluation Metrics.}
To quantitatively evaluate the performance of manipulated activity localization, we establish a standardized evaluation protocol following video action localization benchmarks~\cite{Heilbron2015ActivityNet} and temporal forgery localization~\cite{Zhang2023UMMAFormer}.
We report the Average Precision (AP) at multiple temporal Intersection-over-Union (tIoU) thresholds of $\{0.75, 0.85, 0.95\}$ to assess both localization accuracy and localization precision.
A prediction is considered correct if its tIoU with any ground-truth manipulated segment exceeds the threshold.
We also report the Average Recall (AR) under varying numbers of proposals~$\text{AN} \in \{1, 5, 10\}$.

\noindent
\textbf{Compared Baselines.}
We consider the following baselines for comparisons:
1)~representative temporal action localization methods, including ActionFormer~\cite{Zhang2022ActionFormer}, DiGIT~\cite{Kim2025DiGIT}.
2)~state-of-the-art temporal forgery approaches UMMAFormer~\cite{Zhang2023UMMAFormer} and our proposed TADiff.
All baseline results are reproduced using their official open-source implementations.

\subsubsection{Intra-Domain and Open-World Performance}
Table~\ref{table_inout_domain} presents quantitative comparisons between TADiff and recent state-of-the-art methods on the temporal forgery localization task.
In the \textit{intra-domain} setting, TADiff consistently outperforms all competing methods across both AP and AR  metrics.
The average AP increases from 70.67\% to 75.05\% (+4.38) and the average AR from 74.31\% to 77.15\% (+2.84).
Notably, at the strictest localization threshold (AP@0.95), TADiff achieves a substantial +9.78 improvement, indicating more precise temporal boundary localization.
These results confirm that the proposed diffusion-based feature regularization effectively enhances sensitivity to low-level visual artifacts.

In the \textit{open-world} evaluation, TADiff maintains strong performance on the unseen commercial model, achieving +5.82 AP and +4.61 AR gains, and an impressive +11.98 improvement at AP@0.95.
Unlike typical domain shifts, the performance of all localization methods does not drop compared to the intra-domain setting, thanks to the diverse manipulation mechanisms covered in ActivityForensics that expose localization models to a broad spectrum of domains during training.
This demonstrates that our dataset effectively generalizes to real-world manipulation.
%

\subsubsection{Cross-Domain Generalization}
%
To further evaluate generalization across different manipulation mechanisms, we conduct cross-domain transfer experiments as shown in Table~\ref{table_A_B}.
1)~In the A$\rightarrow$B transfer, TADiff achieves the best performance across all metrics.
The average AP improves from 67.18\% to 69.63\% (+2.45) and the average AR from 72.14\% to 74.91\% (+2.77), with an additional +3.63 gain at AP@0.85.
These results indicate stable boundary localization across heterogeneous forgery mechanisms.
We note that the improvement in AP@0.85 is more significant than in AP@0.95, which reflects the increased difficulty of localizing highly precise manipulation boundaries at a high tIoU threshold of 0.95 under the cross-domain setting.
2)~The B$\rightarrow$A transfer is more challenging, as models require to generalize from a smaller set of simpler generation mechanisms to the more diverse  ones in set A.
Despite this difficulty, TADiff still achieves consistent improvements of +3.75 average AP and +1.53 average AR over the baseline, showing strong robustness to mechanism shifts.
Nevertheless, noticeable performance gaps remain between intra-domain and cross-domain settings, highlighting the inherent challenge of temporal forgery localization across varying manipulation mechanisms.

\subsection{Ablation Studies}
We conduct ablation experiments on ActivityForensics to evaluate the effectiveness of each component in TADiff.

\begin{table}[t!]
\centering
\caption{
Module ablation studies under the intra-domain scenario.
}
\label{table_step}
\scalebox{0.97}{
\begin{tabular}{lccccccccc}
\toprule
Domain
& noise
& denoise
& avg AP
& avg AR
\\
\midrule
\multirow{4}{*}{Intra-Domain}
&\xmark &\xmark
& 70.67 & 74.31
\\
&\cmark &\xmark
& 70.38 & 74.01
\\
&\xmark &\cmark
& 73.52 & 76.22
\\
&\cmark &\cmark
& \textbf{75.05} & \textbf{77.15}
\\
\midrule
\multirow{4}{*}{Open-World}
&\xmark &\xmark
& 77.82 & 83.31
\\
&\cmark &\xmark
& 79.75 & 84.82
\\
&\xmark &\cmark
& 80.10 & 85.58
\\
&\cmark &\cmark
&\textbf{83.64} & \textbf{87.92}
\\
\bottomrule
\end{tabular}}
\end{table}

\begin{figure}[t!]
\centering

\begin{subfigure}{0.9\linewidth}
    \centering
    \includegraphics[width=\linewidth]{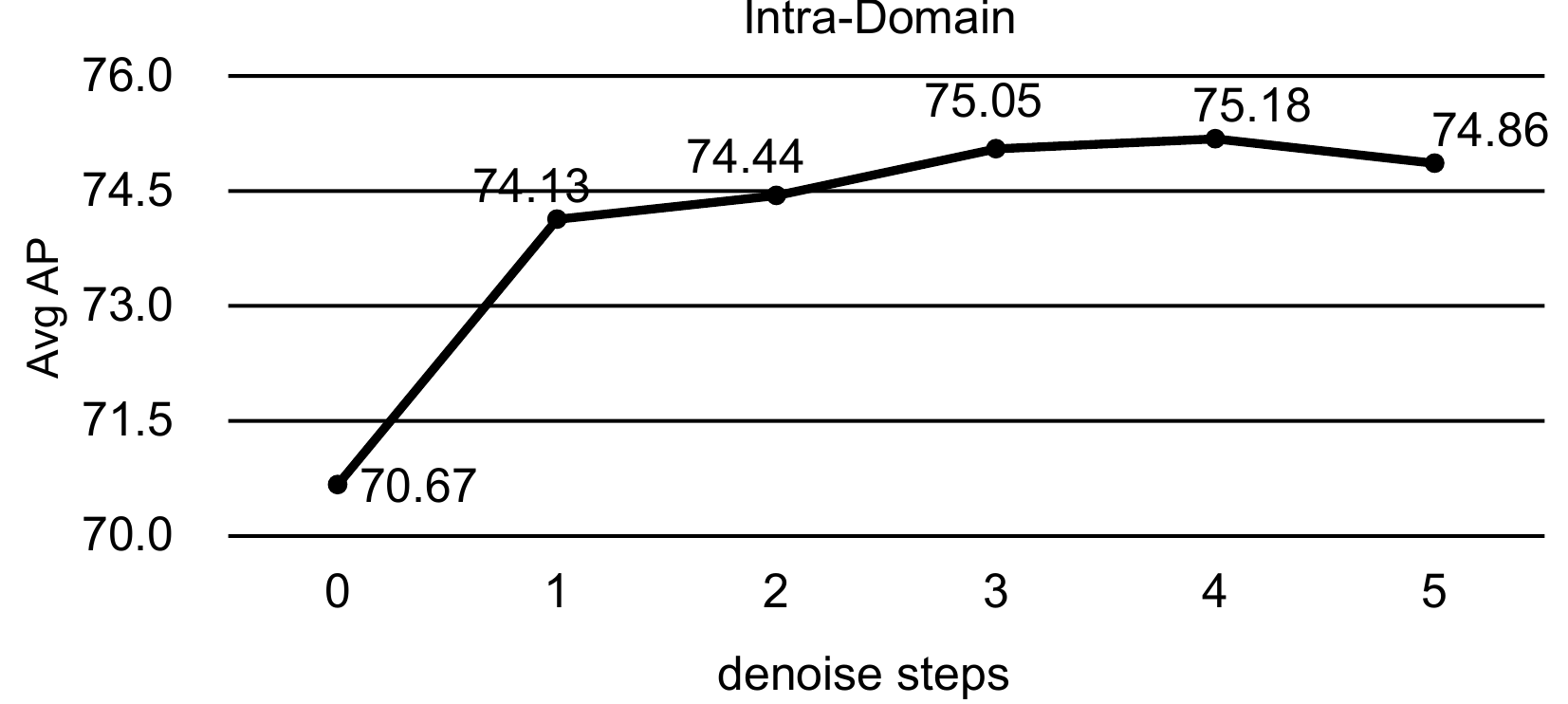}
    \label{fig_step_intra}
\end{subfigure}
\begin{subfigure}{0.9\linewidth}
\vspace{-0.15cm}
    \centering
    \includegraphics[width=\linewidth]{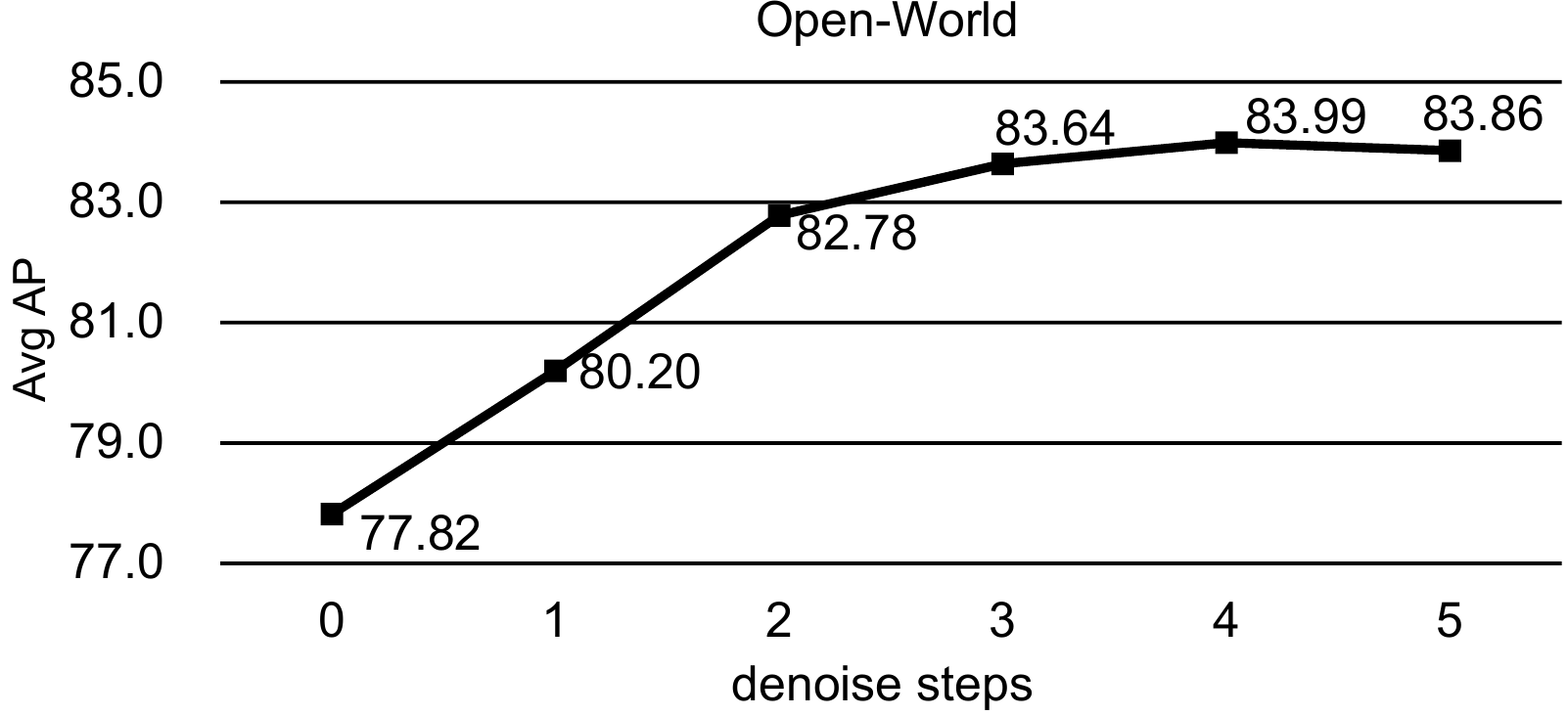}
    \label{fig_step_cross}
\end{subfigure}
\vspace{-0.5cm}
\caption{Impact of denoising step number.}
\label{fig_step_combined}
\vspace{-0.2cm}
\end{figure}

\noindent
\textbf{Module Effectiveness.}
Table~\ref{table_step} reports the results under intra-domain and open-world settings, respectively.
The columns ``noise'' and ``denoise'' indicate whether the forward noise injection module and the reverse denoising module are enabled.
Activating both corresponds to the complete TADiff configuration.
1)~\textit{Noise injection only.}
In the intra-domain setting, performance slightly decreases, indicating that random perturbation may destabilize discriminative features when the training and test distributions are consistent.
In contrast, in the open-world scenario, the same module brings a noticeable improvement (+1.93\% AP), suggesting that noise injection helps break semantic coupling and alleviates over-reliance on content semantics.
2)~\textit{Denoising only.}
This configuration consistently improves performance in both settings, demonstrating that the denoising process enhances temporal structure modeling and feature consistency.
However, it still falls short of the complete configuration, implying that the two modules are complementary: noise injection pushes the model away from the semantic-biased feature space, while denoising reconstructs artifact-sensitive temporal representations.
3)~\textit{Full TADiff (noise + denoise).}
The combination achieves the best results: the average AP/AR rises to 75.05/77.15 under the intra-domain setting and to 83.64/87.92 under the open-world setting.

\noindent
\textbf{Effect of Denoising Steps.}
Fig~\ref{fig_step_combined} shows the effect of different denoising steps $S$ on model performance.
In the intra-domain setting, performance rapidly increases from 0 to 3 steps and peaks at $S=3$ (75.05\% AP), after which it slightly declines, suggesting that only a few iterations are sufficient to recover temporal consistency.
In the open-world setting, the improvement is smoother and the peak appears later ($S=4$, 83.99\% AP), indicating that when the test videos are generated by unseen or commercial models, a longer denoising process helps adapt to distributional discrepancies.

\begin{figure}[t!]
\centering
\includegraphics[width=\linewidth]{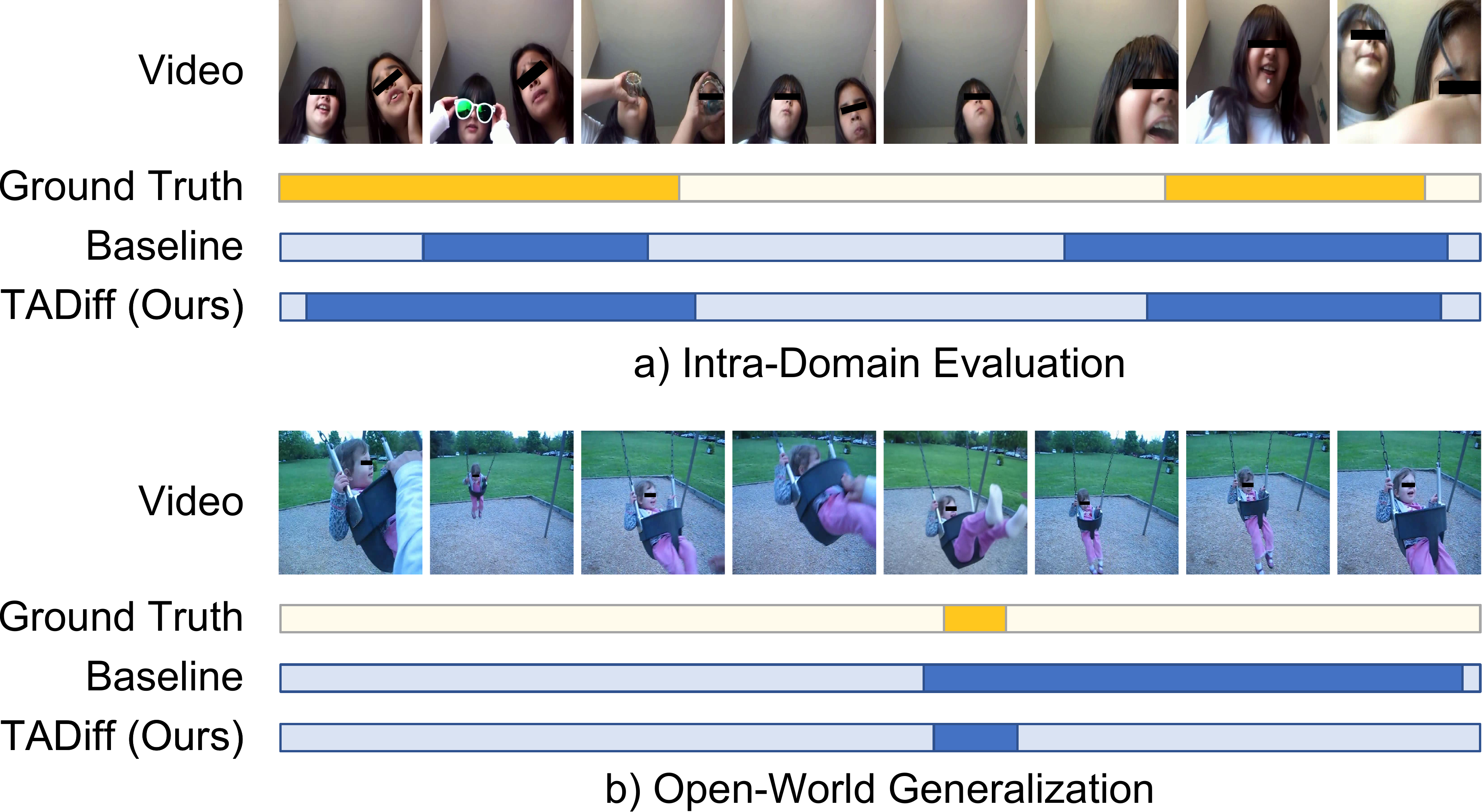}
\caption{
Qualitative comparison of the baseline and TADiff.
The darker yellow rectangle represents the ground-truth forgery segments, while the darker blue one denotes the model's prediction.
}
\label{fig_qualitive}
\end{figure}

\subsection{Qualitative Analysis}
\textbf{Qualitative Comparisons.}
Fig.~\ref{fig_qualitive} presents qualitative comparisons between TADiff and ActionFormer~\cite{Zhang2022ActionFormer} for the temporal forgery localization task, where  TADiff is built upon the ActionFormer architecture.
The upper part shows the intra-domain scenario, and the lower part corresponds to the open-world setting.
In the \textit{intra-domain} case a), ActionFormer can roughly locate the manipulated segments but often suffers from inaccurate temporal boundaries or incomplete coverage.
In contrast, TADiff achieves much tighter alignment with the ground truth intervals, indicating stronger temporal precision under known data distributions.
In the more challenging \textit{open-world} case b), where the forged videos are generated by unseen commercial models, ActionFormer tends to drift or mis-detect authentic regions.
TADiff, however, still accurately captures the manipulated temporal spans, demonstrating better adaptability and robustness to unseen forgery paradigms by effectively reducing semantic bias and improving artifact sensitivity.

\begin{figure}[t!]
\centering
\begin{minipage}{0.495\linewidth}
    \centering
    \includegraphics[width=\linewidth]{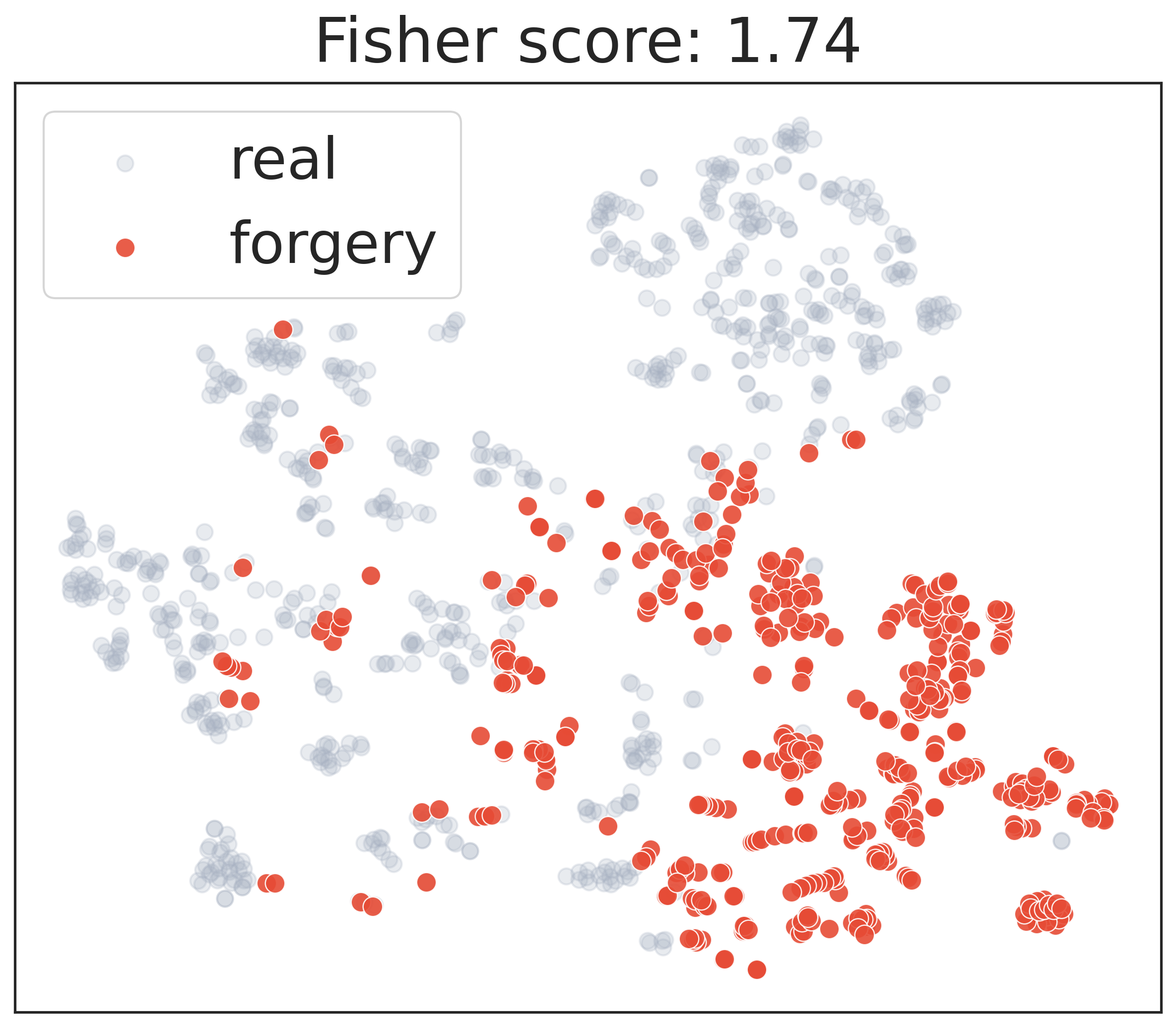}
    \subcaption{
    w/t TADiff 
    (\textit{i.e.,} ActionFormer~\cite{Zhang2022ActionFormer})
    }
\end{minipage}
\hfill
\begin{minipage}{0.495\linewidth}
    \centering
    \includegraphics[width=\linewidth]{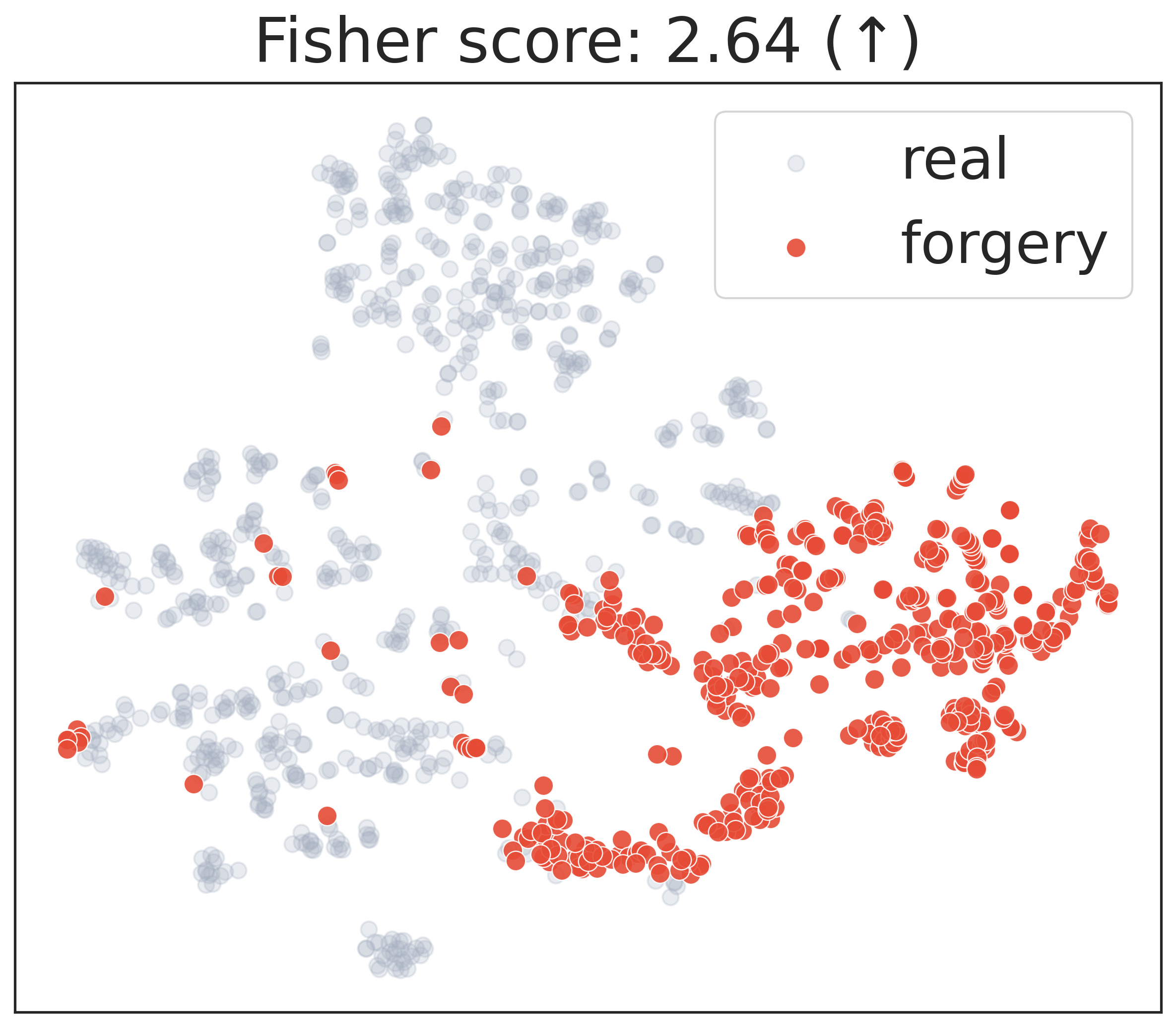}
    \subcaption{
    w/ TADiff 
    }
\end{minipage}
\caption{
t-SNE visualization of features without and with Temporal Artifact Diffuser (TADiff).
The Fisher discriminant score increases from 1.74 to 2.64 after introducing TADiff, which reflects better inter-class separability and reduced intra-class variance in the learned feature space.
}
\label{fig_tsne}
\end{figure}

\noindent
\textbf{Effect of TADiff on Feature Representation.}
To further validate our motivation that TADiff alleviates semantic bias and enhances the model's sensitivity to subtle forgery artifacts, we visualize the learned feature distributions using t-SNE in Fig.~\ref{fig_tsne}.
The left plot corresponds to ActionFormer~\cite{Zhang2022ActionFormer} without TADiff, while the right plot shows the results after integrating TADiff.
Without TADiff, the features of real and forged segments exhibit substantial overlap, indicating that the learned representations are still heavily influenced by high-level semantic information such as scene content and action category, while showing limited discriminability with respect to low-level temporal artifacts.
This semantic entanglement leads to weak separability between authentic and manipulated samples, resulting in a lower Fisher discriminant score of 1.74.
After introducing TADiff, the feature clusters of real and forged segments become clearly separated, and the Fisher discriminant score increases to 2.64.

\section{Conclusion}
In this work, we tackle the emerging challenge of manipulated activity localization, which has become increasingly critical with the advancement of video generation and editing.
We introduce ActivityForensics, the first large-scale dataset specifically designed for localizing manipulated activities in videos.
We propose Temporal Artifact Diffuser (TADiff), a diffusion-based baseline that suppresses semantic bias and amplifies subtle forgery-discriminative signals.
Extensive experiments demonstrate that ActivityForensics and TADiff together provide a strong foundation for advancing activity-level video forgery localization.

\section*{Acknowledgements}
%
%
This research is supported in part by the National Nature Science Foundation of China (NSFC) under Grant 62502187. 
This research is also supported in part by the Natural Science Foundation of Jiangxi Province of China under Grant 20252BAC240015.
%
%
This research is also supported in part by A*STAR under its OTS Research Programme (Award S24T2TS006). Any opinions, findings and conclusions or recommendations expressed in this material are those of the authors and do not reflect the views of the A*STAR.
{
    \small
    \bibliographystyle{meta/ieeenat_fullname}
    \bibliography{meta/citations.bib}
}

\end{document}